\documentclass[a4paper,conference]{IEEEtran}
\usepackage{cite}
\usepackage{graphicx}
\usepackage{amsmath}
\usepackage{amssymb}
\usepackage{booktabs}
\usepackage{multirow}
\usepackage{multicol}
\usepackage{comment}
\usepackage{arydshln}
\usepackage[table]{xcolor}
\usepackage[linesnumbered,ruled,vlined]{algorithm2e}
\usepackage[pagebackref,breaklinks,colorlinks]{hyperref}
\usepackage{subfig}
\usepackage{flushend}

\def\blue#1{\textcolor[rgb]{0,0,0}{#1}}

\begin{document}

%\title{Self-Supervised Patch-Wise Attention-guided Reconstruction and Dual Triplet Loss for Writer Independent Offline Signature Verification}
\title{SURDS: Self-Supervised Attention-guided Reconstruction and Dual Triplet Loss for Writer Independent Offline Signature Verification}
%\title{ARSE: Attention-guided Reconstruction-based Self-Supervised Representation Learning and Dual Triplet Loss for Writer Independent Offline Signature Verification}

% author names and affiliations
% use a multiple column layout for up to three different
% affiliations
\author{
\IEEEauthorblockN{Soumitri Chattopadhyay}
\IEEEauthorblockA{Dept. of Information Technology\\
Jadavpur University, Kolkata\\
Email: soumitri.chattopadhyay@gmail.com}\\
\IEEEauthorblockN{Saumik Bhattacharya}
\IEEEauthorblockA{Dept. of Electronics and Electrical Communication Engineering\\
Indian Institute of Technology Kharagpur\\
Email: saumik@ece.iitkgp.ac.in}
\and\and
\IEEEauthorblockN{Siladittya Manna}
\IEEEauthorblockA{Computer Vision and Pattern Recognition Unit\\
Indian Statistical Institute, Kolkata\\
Email: siladittya\_r@isical.ac.in}\\
\IEEEauthorblockN{Umapada Pal}
\IEEEauthorblockA{Computer Vision and Pattern Recognition Unit\\
Indian Statistical Institute, Kolkata\\
Email: umapada@isical.ac.in}
}
% make the title area
\maketitle

\begin{abstract}
Offline Signature Verification (OSV) is a fundamental biometric task across various forensic, commercial and legal applications. The underlying task at hand is to carefully model fine-grained features of the signatures to distinguish between genuine and forged ones, which differ only in minute deformities. This makes OSV more challenging compared to other verification problems. In this work, we propose a two-stage deep learning framework that leverages self-supervised representation learning as well as metric learning for writer-independent OSV. First, we train an image reconstruction network using an encoder-decoder architecture that is augmented by a 2D spatial attention mechanism using signature image patches. Next, the trained encoder backbone is fine-tuned with a projector head using a supervised metric learning framework, whose objective is to optimize a novel dual triplet loss by sampling negative samples from both within the same writer class as well as from other writers. The intuition behind this is to ensure that a signature sample lies closer to its positive counterpart compared to negative samples from both intra-writer and cross-writer sets. This results in robust discriminative learning of the embedding space. To the best of our knowledge, this is the first work of using self-supervised learning frameworks for OSV. The proposed two-stage framework has been evaluated on two publicly available offline signature datasets and compared with various state-of-the-art methods. It is noted that the proposed method provided promising results outperforming several existing pieces of work. The code is publicly available at: \href{https://github.com/soumitri2001/SURDS-SSL-OSV}{https://github.com/soumitri2001/SURDS-SSL-OSV}.
\end{abstract}

% no keywords

% For peer review papers, you can put extra information on the cover
% page as needed:
% \ifCLASSOPTIONpeerreview
% \begin{center} \bfseries EDICS Category: 3-BBND \end{center}
% \fi
%
% For peerreview papers, this IEEEtran command inserts a page break and
% creates the second title. It will be ignored for other modes.
\IEEEpeerreviewmaketitle

%%%%%%%%%%%%%%%%%%%%%%%%%%%%%%%%%%%%%%%%%%%%%%%%%%%%%%%%%%%%%%%%%%%%%%%%%%%%%

\section{Introduction}
Handwritten signature is the most common mode of biometric identification that has been used over decades across various aspects of human activity. From bank cheques, legal documents to offline application forms, signature has been the de facto standard method for document validation. Thus, verification of a signature is a very crucial task to accurately assert the original owner of a signature sample, since fraudulence is widespread and often highly skilled. Manual verification is time consuming as well as prone to human error, thereby required to be automated by pattern recognition methods.

\begin{figure}
    \centering
    \includegraphics[width=0.9\columnwidth]{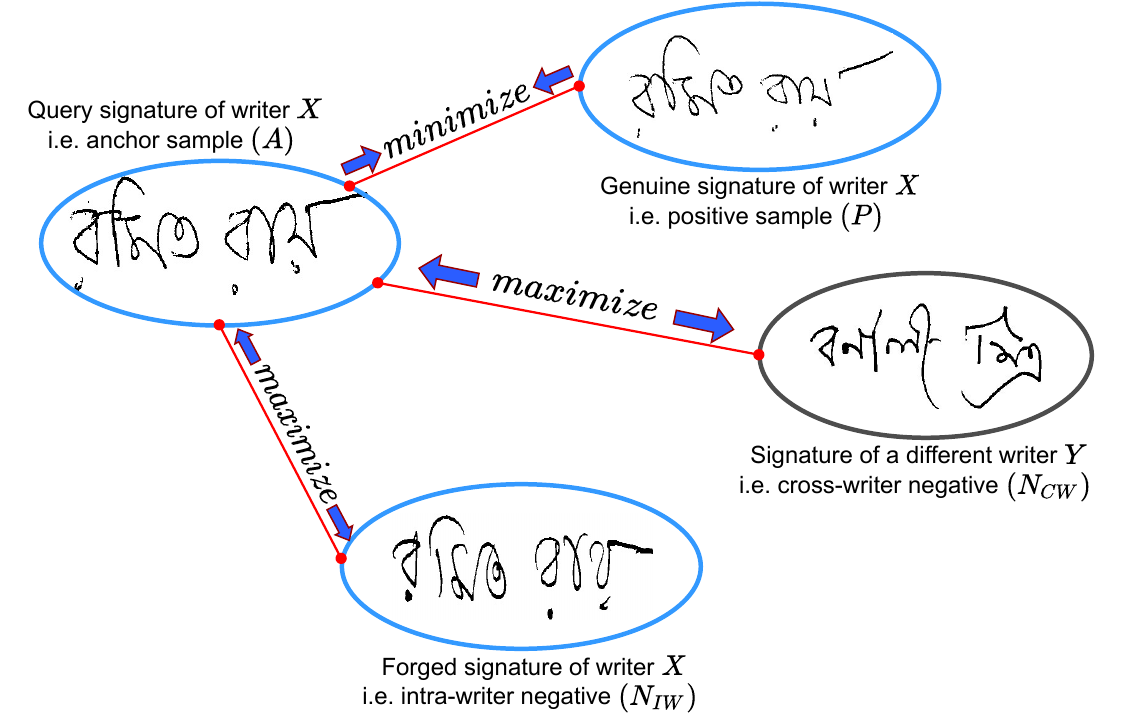}
    \caption{Schematic of the intuition behind using the dual triplet objective to fine-tune the pre-trained self-supervision model. More details have been provided in Section \ref{subsec:dualtriplet}.}
    \label{fig:dtl_intuit}
\end{figure}

Signatures may be categorised as online and offline, depending on the modality it is obtained. Online signatures contain timestep-wise coordinate information of a signature, usually obtained from an electronic writing pad. Since such devices may be associated with sensors that can record several parameters such as pen pressure, time taken per character and so on, which are unique to a particular writer, verifying online signatures is a fairly easy task. On the other hand, offline signatures are 2D images that are captured electronically or manually before being processed for verification. Such images contain no explicit information that can define a writing behaviour, thereby requiring minute modeling of fine-grained features such as character strokes and handwriting style. Thus, purely vision-based signature verification is a challenging task and thus, has been widely studied over the last few decades.

Broadly speaking, two approaches have been devised for OSV, they being (1) writer dependent and (2) writer independent. In case of writer dependent OSV, a classifier is trained for every writer class, which poses scalability problems as they require the system to be re-trained each time a new user is to be registered, as well as requires a large number of signature samples from every user which is not easily available at one go. However, writer independent OSV systems are more focused towards finding intrinsic differences between genuine and forged samples such that they can be easily extended to new users without the need of re-training the system every time. Thus, it is intuitive that the latter scenario is more close to the real-world that can be easily up-scaled to industry standards. 

In this research, we aim to develop a novel framework that leverages self-supervised representation learning followed by supervised fine-tuning for writer independent OSV. Although self-supervision has been employed to various domains of computer vision \cite{jing2020self, jaiswal2021survey, shurrab2021self}, their application to offline handwritten data has been limited. This might be due to the fact that existing self-supervised approaches such as image colorization \cite{zhang2016colorful}, super-resolution \cite{ledig2017photo} or even contrastive learning-based methods \cite{jaiswal2021survey, chen2020simple} are designed for pixel perfect natural or biomedical images \cite{shurrab2021self} and thus, may not be suitable for sparse binarized data \cite{bhunia2021vectorization}. In fact, to the best of our knowledge there has been no prior work on offline signature verification that has employed self-supervision strategy, which has been a driving motivation for our research. 

In particular, we propose an encoder-decoder network for self-supervised image reconstruction that is augmented by a patch-wise 2D attention module. The attention mechanism allows the model to focus on visually important regions in the entire image in a patch-wise manner. The intuition behind this mechanism is to identify fine-grained local information from the signature samples. Next, we incorporate a metric learning approach for fine-tuning the pre-trained encoder by formulating an objective that separates negative samples from both within the same writer as well as samples from different writers. Specifically, we use a dual triplet loss module that is computed using representations from both intra-writer and cross-writer negative samples with respect to an anchor-positive pair of intra-writer signatures. The cross-writer samples are implicitly negative samples as they come from different writers. The intuition has been schematically depicted in \autoref{fig:dtl_intuit}. This allows a superior separation of genuine and forgeries, which has been empirically shown while comparing with methods that use a single metric learning objective \cite{dey2017signet}.  Finally, the frozen encoder network is used as a feature extractor for evaluation purpose. We have evaluated the proposed model on two publicly available signature datasets \cite{pal2016performance}, where it has yielded results that are better or at par with other state-of-the-art works in literature.

To summarize, the main contributions of the presented work are as follows:

\begin{itemize}
    \item A novel two-stage framework, named SURDS, has been proposed for writer independent OSV. 
    %\item A self-supervised attention-guided reconstruction-based pre-training strategy has been proposed for OSV, after which a dual triplet loss-based fine-tuning is employed to tackle both intra-writer and cross-writer negative samples.
    \item A self-supervised attention-guided reconstruction-based pre-training strategy has been proposed for learning fine-grained representations from signature images. 
    \item A dual triplet loss-based fine-tuning is also proposed to tackle both intra-writer and cross-writer negative samples.
    \item To the best of our knowledge, this is the first work to employ self supervised learning concept in the context of OSV.
    \item The proposed framework outperforms adapted baseline competitors and also surpasses several state-of-the-art works on two publicly available offine signature datasets. 
\end{itemize}

The rest of the paper is organized as follows: Section \ref{sec:litrev} discusses some of the works in literature pertaining to OSV and self-supervised learning approaches; Section \ref{sec:method} presents a detailed explanation of the proposed framework for signature verification from offline images; Section \ref{sec:exptdet} describes the implementation procedure and experimental protocol used in this work; Section \ref{sec:results} analyses the empirical performance of the proposed model against adapted baselines and state-of-the-art works; and finally, Section \ref{sec:concl} concludes the findings of the proposed research.

\section{Related Works}\label{sec:litrev}

\subsection{Offline Signature Verification}
OSV has been well studied in recent decades, with the surveys \cite{hafemann2017offline, kaur2021signature} providing comprehensive review of the literature. Although classically approached using handcrafted feature extraction and analysis \cite{pal2016performance, dutta2016compact, alaei2017efficient}, deep learning-based methods have also been introduced in recent years, such as the works presented in \cite{dey2017signet} that introduced contrastive loss-based Siamese CNNs for signature verification; \cite{shariatmadari2019patch} which proposed a hierarchical CNN for learning features from signature patches; and \cite{berkay2018hybrid} that employed a hybrid two-channel CNN for OSV. More recent works include region-based deep metric learning \cite{liu2021offline}, neuromotor inspired framework \cite{diaz2016approaching}, Siamese networks followed by support vector classifier \cite{rateria2018off}, capsule networks-based model \cite{parcham2021cbcapsnet} and a graph neural network approach \cite{roy2021offline}. Typically, OSV has been approached using fully supervised methods only. In this work, we try to develop a self-supervised pre-training method for writer independent signature verification.

\subsection{Self-Supervised Learning}
Self-supervised learning \cite{jing2020self, jaiswal2021survey} has emerged as a powerful paradigm that seeks to learn meaningful representations from unlabelled data to recognise underlying patterns without the need for explicit annotation. This largely alleviates the problem of scarcity of labelled data for specialised tasks, since the unsupervised pre-trained representations are transferable to the downstream task at hand. Classical means of self-supervised learning include generative modelling \cite{baldi2012autoencoders, goodfellow2014generative}, similarity \cite{chen2021exploring, zbontar2021barlow} and contrastive learning approaches \cite{chen2020simple, he2020momentum}. Particularly in the context of computer vision, self-supervision has been explored by several pretext tasks for representation learning such as image colorization \cite{zhang2016colorful}, inpainting \cite{pathak2016context}, super-resolution \cite{ledig2017photo} and solving jigsaw puzzles \cite{noroozi2016unsupervised}, among others. However, such methods are designed for natural images and hence are not suitable for sparse grayscale signatures. Bhunia et al. \cite{bhunia2021vectorization} employed cross-modal translation  between vector and raster space for online co-ordinate based handwriting recognition. %which however is unsuitable for our purpose since it requires the availability of paired offline images and online coordinates. 
To this end, we propose a first-of-its-kind self-supervised pre-training strategy for OSV that uses a patch-wise attention guided reconstruction network for learning representations from signature images.

\section{Methodology}\label{sec:method}

\subsection{Preprocessing of Signature Images}\label{subsec:preproc}
\begin{figure}
    \centering
    \includegraphics[width=\columnwidth]{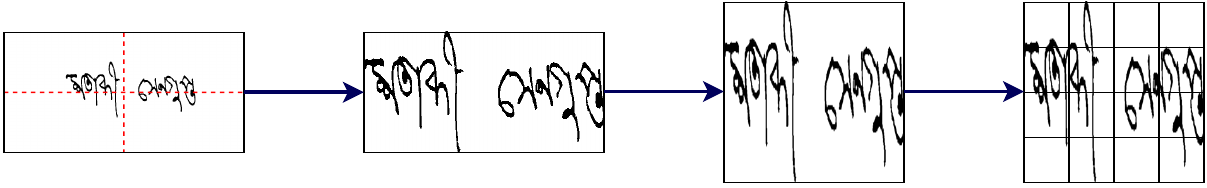}
    \caption{Illustration of the preprocessing steps applied on the raw signature images before feeding them into the encoder network.}
    \label{fig:preproc}
\end{figure}

Since raw signature images contain a lot of redundant background pixels, instead of directly dividing them into patches, we devise a preprocessing technique to ensure that the signatures are cropped to a tight bound so as to minimize background pixels. We do so by first computing the center of mass of the raw binarized image, after which we traverse through vertical and horizontal axes on either side of the center of mass coordinates to find the location of the last foreground pixel, at which we crop the image. Once cropped to a tight bound, the images are resized to $256\times256$ pixels and normalized before being fed into the encoder network. Further, this resized image is divided into non-overlapping patches of $64\times64$ pixels, thereby obtaining $16$ patches for each image. A figure illustrating the process has been shown in \autoref{fig:preproc}.

\subsection{Self-Supervised Pre-training}\label{subsec:ssl_pretr}

\begin{figure*}
    \centering
    \includegraphics[width=0.8\linewidth]{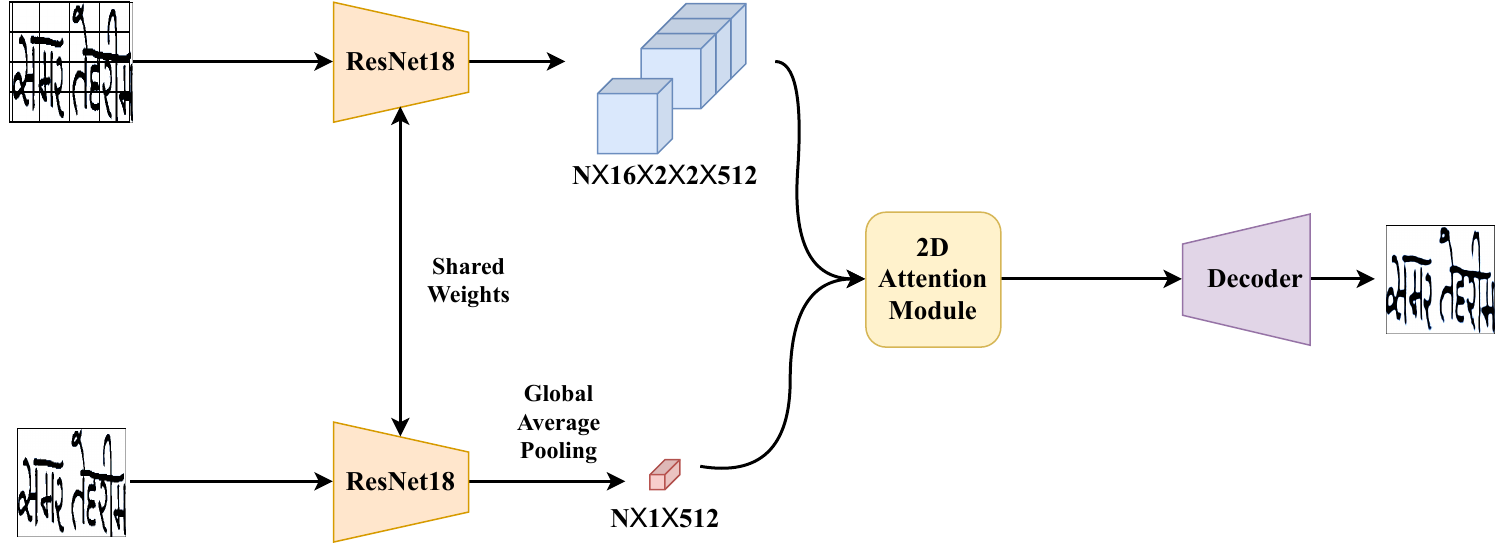}
    \caption{Graphical representation of the proposed self-supervised reconstruction-based pre-training pipeline for offline signature verification.}
    \label{fig:sslarch}
\end{figure*}

The goal of any self-supervised pipeline \cite{zhang2016colorful, pathak2016context} is to design a pre-training strategy by which a model learns meaningful representations that can be used off-the-shelf or fine-tuned for a downstream task. For signature images, we set the goal to learn representations that contain local region level information. To achieve this goal, we have designed a patch-wise attention-guided image reconstruction network, as shown in \autoref{fig:sslarch}. Thus, the pipeline comprises of an encoder-decoder architecture, the details of which have been put forth in the following section.

\subsubsection{Convolutional Encoder-Decoder Architecture}\label{subsubsec:enc_dec}
For the convolutional encoder, we used the ResNet-18 \cite{he2016deep} model, which takes in a signature image $I \in \mathcal{R}^{H \times W \times 3}$ and outputs a convolutional feature map $\mathcal{C} \in \mathcal{R}^{h \times w \times 512}$ having $512$ channels. While feeding the full image into the network, we perform a global average pooling on the obtained feature map to get a feature vector of dimension $512$, whereas the pooling operation is absent when the signature patches are fed. This is due to the fact that we intend to perform a 2D attention between the patch feature maps and the global feature vector, the details of which have been put in the subsequent section. 

For the decoder network, we have used the state-of-the-art U-Net \cite{ronneberger2015unet} decoder architecture, excluding the skip connections. The decoder takes the attention-enriched $512$-dimensional feature vector as input and performs a series of upsampling operations comprising transpose convolutions to finally output a tensor $Y \in \mathcal{R}^{H \times W \times 3}$ i.e. the dimensions of the original signature image $I$. %We have excluded the skip connections from the encoder which are present in the original paper \cite{ronneberger2015unet}, where the authors aimed to perform semantic segmentation. 
We formulate the reconstruction loss as the simple mean squared error between the original and reconstructed images, which thereby constitutes the self-supervision task objective given by \autoref{eq:ssl}. 

\begin{equation}\label{eq:ssl}
    \mathcal{L}_{SSL} = \frac{1}{H \times W} \sum_{i=1}^{H} \sum_{j=1}^{W} \lvert I_{ij} - Y_{ij} \rvert^2
\end{equation}

\subsubsection{Patch-wise 2D Attention Mechanism}\label{subsubsec:pw_attn}

\begin{figure*}
    \centering
    \includegraphics[width=0.8\linewidth]{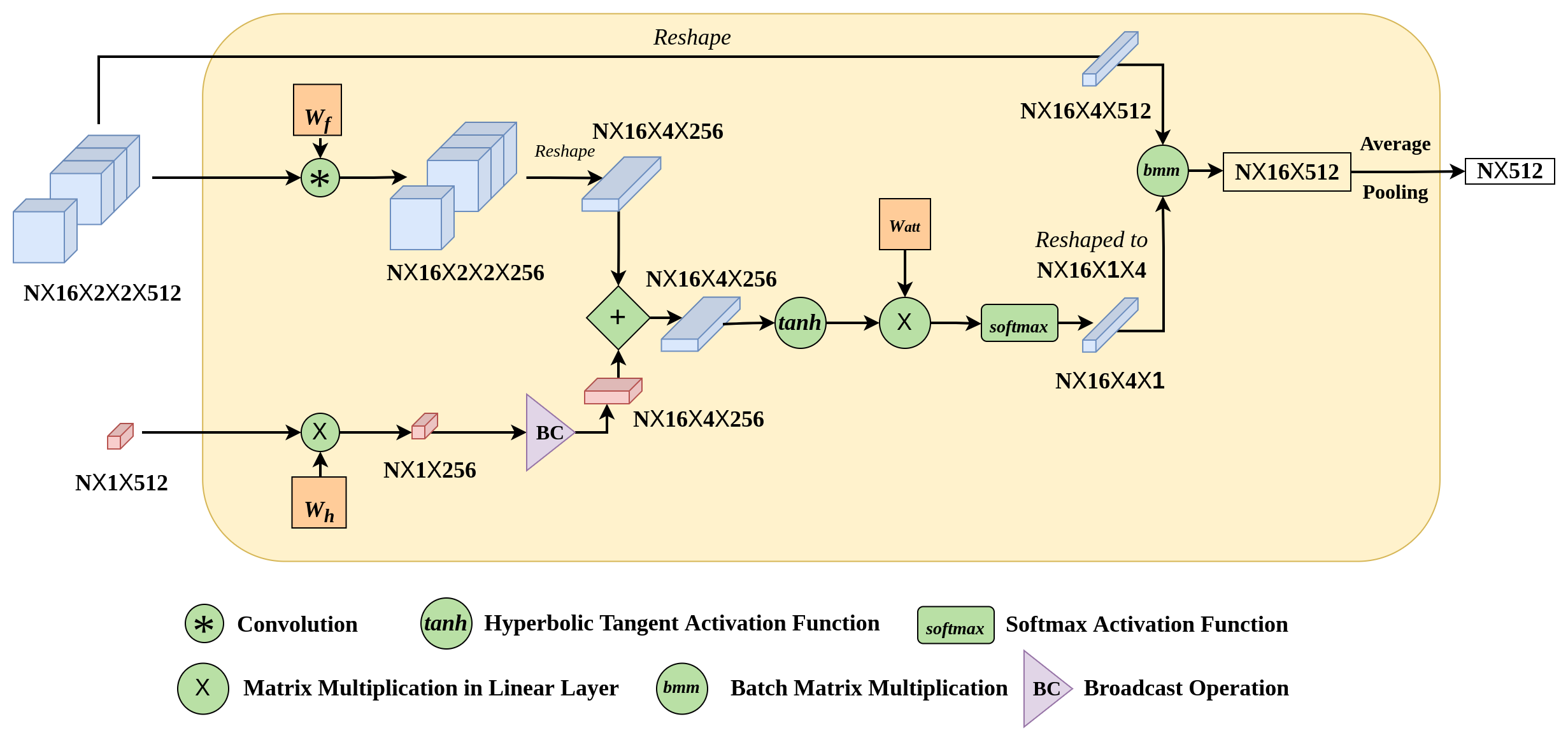}
    \caption{Schematic diagram showing the 2D attention mechanism employed in the proposed study. The ``broadcast operation'' refers to that of PyTorch \cite{paszke2019pytorch} framework by which tensors of smaller dimensions are vectorized for mathematical operations with a bigger dimensional tensor.}
    \label{fig:attn}
\end{figure*}

This is the key module in the self-supervised training pipeline that enforces focusing on visually important parts of the signature in a region-wise fashion. The diagrammatic representation of this module is shown in \autoref{fig:attn}. We use the convolution feature maps $\mathcal{C}$ obtained from the signature patches and the feature vector $\mathcal{Z}$ obtained from the image $I$ using the same weight shared encoder. The module outputs an attention enriched context vector $\mathcal{H}$ using the equations given below:

\begin{equation}
\begin{split}
    & \mathcal{M} = tanh(\mathcal{W}_f \circledast \mathcal{C} + \mathcal{W}_h \mathcal{Z}) \\
    & \mathcal{A}_{ij} = softmax(\mathcal{W}_{att} \mathcal{M}_{ij})  \\
    % & \textcolor{black}{\mathcal{H} = \sum_{i=1}^{h} \sum_{j=1}^{w} \mathcal{A}_{ij} \cdot \mathcal{C}_{ij} }\\
    & \textcolor{black}{\mathcal{H}_{ij} = \sum_{k} \mathcal{A}^T_{ik} \cdot \mathcal{C}_{kj} }
\end{split}
\end{equation}

Here, $\mathcal{W}_f$, $\mathcal{W}_h$ and $\mathcal{W}_{att}$ are learnable weights, while the ``$\circledast$'' denotes the convolution operation. The attention weight at $(i,j)$ is computed by $\mathcal{A}_{ij}$, while $\mathcal{H}$ denotes the attention enriched context feature vector obtained. For every patch feature map $\mathcal{C}_k$ we obtain a context vector $\mathcal{H}_k$, which is then accumulated and a global average pooling operation is performed to obtain a single feature vector of $512$ dimensions. This vector is then passed into the decoder for reconstruction.

\subsection{Metric Learning using Dual Triplet Loss}\label{subsec:dualtriplet}
Once the self-supervised pre-training is complete, we take the pre-trained ResNet-18 encoder and attach a projector head comprising two embedding layers of $512$ dimensions each. This network is then fine-tuned using a metric learning objective that distances a given query signature from both intra-writer and cross-writer negatives, while bringing a positive signature closer to itself (please refer to \autoref{fig:dtl_intuit} for a graphical form of the described intuition). To do so, we have used the triplet loss function, a widely used loss objective in deep metric learning that takes the representation of a query (anchor, $A$) sample and minimizes its distance with a positive class sample ($P$) and simultaneously maximizes its distance with a negative class sample ($N$). The individual losses have been defined in \autoref{eq:lossfn}.

\begin{equation}\label{eq:lossfn}
\begin{split}
    & \mathcal{L}_{IW} = \max \{ 0, \delta(z_A, z_P) - \delta(z_A, z_{N_{IW}}) + \mu \} \\
    & \mathcal{L}_{CW} = \max \{ 0, \delta(z_A, z_P) - \delta(z_A, z_{N_{CW}}) + \mu \}
\end{split}
\end{equation}

Where, $\mu > 0$ is a hyperparameter used to avoid trivial solution, while $\delta(z_i, z_j) = \lVert z_i - z_j \rVert_{2}$.

Combining the intra-writer and cross-writer losses, we define our dual triplet objective as:

\begin{equation}
    \mathcal{L}_{DT} = \mathcal{L}_{IW} + \mathcal{L}_{CW}
\end{equation}

% \begin{figure}
%     \centering
%     \includegraphics[width=\columnwidth]{triplet_intuition.pdf}
%     \caption{Schematic of the intuition behind using the dual triplet objective in our proposed approach for fine-tuning the previously trained self-supervised model.}
%     \label{fig:dtl_intuit}
% \end{figure}

\subsection{Downstream Evaluation}\label{subsec:dstream}
After the pre-training phase is completed, we take the frozen backbone comprising the encoder and projector and use it for feature extraction from unseen signature images for downstream evaluation. We have used the evaluation protocol described by Dey et al. \cite{dey2017signet} for writer independent signature verification. Specifically, we have used a threshold $\tau$ that determines if a pair of signature $(i, j)$ is genuine or forged depending on a distance metric $\mathcal{D}(z_i, z_j)$ between their embeddings, as given by \autoref{eq:dist}. However, genuineness/falsification is decided using a ground truth reference. For this purpose, we randomly select $8$ genuine samples from each of the unseen writers and consider them as reference signatures, and the mean of the distance metric between each of the references with the queried sample is considered for comparison with the threshold. The remaining samples are then used for evaluation.  

\begin{equation}\label{eq:dist}
    \mathcal{D}(z_i, z_j) = \lVert z_i - z_j \rVert
\end{equation}

\section{Experiments}\label{sec:exptdet}

\subsection{Datasets}\label{subsec:dataset}
We have used two publicly available offline signature datasets, BHSig260 Bengali and Hindi \cite{pal2016performance} for evaluating our proposed two-stage pipeline. The Bengali signature dataset comprises signatures of 100 writers, while the Hindi dataset consists of signatures of 160 writers. For each writer, there are with 24 genuine signature samples and 30 skilled forgeries. We have split the datasets randomly in 0.7/0.3 ratio for training/testing i.e. 70/30 writers for the Bengali and 112/48 writers for the Hindi datasets, respectively. Since our premise is writer independent signature verification, we have ensured that the writers \emph{do not} overlap between the respective sets.

% \begin{table}[!ht]
% \centering
% \caption{Number of writers in train and test splits of the datasets used to evaluate the proposed model. \label{tab:dset_split}}
% % \resizebox{\columnwidth}{!}{
% \begin{tabular}{c|c|c} 
% \hline
% \multirow{2}{*}{\textbf{Dataset}} & \multicolumn{2}{c|}{\textbf{No. of writers}}        \\ 
% \cline{2-3}
% & \multicolumn{1}{c}{\textbf{Train}} & \textbf{Test}  \\ \hline
% BHSig260 Bengali  & 70    & 30             \\
% BHSig260 Hindi    & 112   & 48            
% \end{tabular}
% % }
% \end{table}

\subsection{Implementation}\label{subsec:impl}
We have implemented our proposed framework using the PyTorch framework \cite{paszke2019pytorch} on a 8GB Nvidia GeForce RTX 2080 GPU. In the self-supervised pre-training phase, a randomly initialized ResNet-18 \cite{he2016deep} excluding linear layers is used as the encoder, trained with SGD \cite{sutskever2013importance} optimizer with initial learning rate of $0.1$. The pre-trained encoder is then used in the fine-tuning phase, where the network is trained using RMSProp with initial learning rates for the encoder and the projector being $5.10^{-3}$ and $1.0$, respectively and weight decay factor of $5.10^{-4}$. Both training phases are conducted for a maximum of 200 epochs with a cosine annealing learning rate scheduler. Prior to being fed into the encoder, the signatures have been preprocessed using the steps described in Section \ref{subsec:preproc}.

% We have implemented our proposed framework using the PyTorch framework \cite{paszke2019pytorch} on a 8GB Nvidia GeForce RTX 2080 GPU. For the CNN encoder used in self-supervised pre-training, we have used a randomly initialized ResNet-18 \cite{he2016deep} network excluding the linear layers, which is then re-used off-the-shelf for the metric learning phase after pre-training is completed. The optimizer used for the pre-training task is SGD \cite{sutskever2013importance} with an initial learning rate of $0.1$. For the fine-tuning phase, we have used the RMSprop optimizer with initial learning rates for the encoder and the projector being $0.005$ and $1.0$ respectively and weight decay factor of $0.0005$. For both the phases, the respective models have been trained for 200 epochs, and a Linear Warmup Cosine Annealing scheduler has been used for learning rate decay. Prior to being fed into the encoder, the signature images have been preprocessed using the steps described in Section \ref{subsec:preproc}.

\subsection{Evaluation Metrics}\label{subsec:evalmet}

For signature pairs $(i,j)$, let $\mathcal{G}$ and $\mathcal{G'}$ denote sets of genuine-genuine and genuine-forged signature pairs respectively. Then, at a given threshold $\tau$ we can define the set of true positives ($T_P$), true negatives ($T_N$), false positives ($F_P$) and false negatives ($F_N$) as:

% \begin{table}[]
%     \centering
%     \begin{tabular}{c|c|c}
%          & $\mathcal{D}(z_i, z_j) \leq \tau$ & $\mathcal{D}(z_i, z_j) > \tau$ \\ \hline \hline
%         $(i,j) \in \mathcal{G}$ & $T_P(\tau)$ & $F_N(\tau)$ \\ \hline
%         $(i,j) \in \mathcal{G'}$ & $F_P(\tau)$ & $T_N(\tau)$\\
%     \end{tabular}
%     \caption{Caption}
%     \label{tab:my_label}
% \end{table}

\begin{equation}
\begin{split}
    & T_P(\tau) = \{(i,j) \in \mathcal{G}, \mathcal{D}(z_i, z_j) \leq \tau \} \\
    & T_N(\tau) = \{(i,j) \in \mathcal{G'}, \mathcal{D}(z_i, z_j) > \tau \} \\
    & F_P(\tau) = \{(i,j) \in \mathcal{G'}, \mathcal{D}(z_i, z_j) \leq \tau \} \\
    & F_N(\tau) = \{(i,j) \in \mathcal{G}, \mathcal{D}(z_i, z_j) > \tau \}
\end{split}
\end{equation}

Accordingly, the respective rates may be defined as the ratio of set cardinality to its superset cardinality, denoted by $n(\cdot)$.

\begin{equation}
\begin{split}
    & TPR(\tau) = \frac{n(T_P(\tau))}{n(\mathcal{G})} ;\;\;\;\; TNR(\tau) = \frac{n(T_N(\tau))}{n(\mathcal{G'})} \\
    & FPR(\tau) = \frac{n(F_P(\tau))}{n(\mathcal{G'})} ;\;\;\;\; FNR(\tau) = \frac{n(F_N(\tau))}{n(\mathcal{G})}
\end{split}
\end{equation}

Based on these, we define the three evaluation metrics used in this work:

\begin{itemize}

\item \textbf{Accuracy (ACC): }
The accuracy is defined as the arithmetic mean of the true positive and true negative rates. We report the maximum accuracy achieved by varying the threshold $\tau$ from the minimum distance to maximum distance value at intervals of $5e-5$.

\begin{equation}
    Accuracy = \max\limits_{\tau \in \mathcal{D}} \left\{ \frac{TPR(\tau) + TNR(\tau)}{2} \right\}
\end{equation}

\item \textbf{False Acceptance Rate (FAR): }
FAR is defined as the ratio of forged samples incorrectly determined to be genuine and thereby accepted (i.e. $FPR$), 

\item \textbf{False Rejection Rate (FRR): }
FRR is defined as the ratio of genuine samples incorrectly predicted to be forgeries and thereby rejected (i.e. $FNR$).
\end{itemize}

\section{Results and Analysis}\label{sec:results}

\subsection{Ablation Study}\label{subsec:comp_base}
We first compare our proposed framework with some ablated baseline models derived from our architecture:

\begin{itemize}
    \item \textbf{RN-DTL: } We ablate the self-supervised pre-training process and train a randomly initialized ResNet-18 encoder using the dual triplet loss based supervised metric learning proposed in our study.
    
    \item \textbf{AE-DTL: } We ablate the attention mechanism and simply train an  autoencoder using the encoder-decoder architecture described in Section \ref{subsubsec:enc_dec}, followed by fine-tuning of the encoder using our proposed dual triplet loss. 
\end{itemize}   

    % \item \textbf{SimCLR: } We adopt a state-of-the-art self-supervised algorithm, SimCLR \cite{chen2020simple} and train it end-to-end on the respective datasets. Thus, this forms a fully self-supervised baseline model.

The results of the comparative study have been tabulated in \autoref{tab:comp_ablative}. From the results, it is evident that self-supervised pre-training significantly boosts the downstream signature verification performance. %\textcolor{black}{Further, that the AE-DTL model comes very close to that of the proposed method on the Hindi dataset \cite{pal2016performance} in terms of performance highlights the usefulness of a reconstruction-based pre-training approach for OSV.}

\begin{table}[!ht]
\centering
\caption{Results obtained from the ablation study. \label{tab:comp_ablative}}
\resizebox{\columnwidth}{!}{
\begin{tabular}{c|ccc|ccc} 
\hline
\multirow{2}{*}{\textbf{Baseline}} 
& \multicolumn{3}{c|}{\textbf{BHSig260 Bengali} \cite{pal2016performance}}                 
& \multicolumn{3}{c}{\textbf{BHSig260 Hindi} \cite{pal2016performance}}                     
\\ \cline{2-7}
& \textbf{Accuracy (\%)} & \textbf{FAR (\%)} & \textbf{FRR (\%)} 
& \textbf{Accuracy (\%)} & \textbf{FAR (\%)} & \textbf{FRR (\%)}  
\\ \hline
RN-DTL                              
& 82.18                  & 21.66             & 13.95             
& 86.72                  & 14.67             & 11.88              \\
AE-DTL                              
& 83.57                  & 29.11             & 3.75              
& 89.14                  & 18.47             & 3.25               \\
% SimCLR \cite{chen2020simple}                            
% & 73.45                  & 11.71             & 54.30             
% & 72.45                  & 10.34             & 59.91              \\ 
\hline
\textbf{Proposed Method}           
& \textbf{87.34}         & \textbf{19.89}    & \textbf{5.42}     
& \textbf{89.50}         & \textbf{12.01}    & \textbf{8.98}      \\
\hline
\end{tabular}
}
\end{table}

% ROC -- trial
% \begin{figure}[b]
%     \centering
%     \subfloat[BHSig260 Bengali]{\includegraphics[width=115pt, height=115pt]{ROC_87acc_B3.png}}\;
%     \subfloat[BHSig260 Hindi]{\includegraphics[width=115pt, height=115pt]{ROC_87acc_B3.png}}\;
%     \caption{ROC}
%     \label{fig:roc}
% \end{figure}

\subsection{Comparison with state-of-the-art}\label{subsec:comp_sota}
\blue{To the best of our knowledge, there is no prior work that uses self-supervised pre-training for writer independent OSV. Thus, we adopt some of the recent state-of-the-art self-supervised algorithms such as SimCLR} \cite{chen2020simple}, \blue{Barlow Twins} \cite{zbontar2021barlow} \blue{and SimSiam} \cite{chen2021exploring} \blue{for pre-training, followed by downstream fine-tuning using the proposed dual triplet loss. Further, we also compare our proposed framework with some of the existing state-of-the-art supervised methods by various researchers in literature. The empirical results are shown in \autoref{tab:comp_sota}. It is evident that our proposed pipeline shows promising performance on both the datasets, outperforming popular self-supervision algorithms as well as several of the existing works by large margins. Furthermore, the FRR values obtained by the proposed method are significantly better than all of the existing works.  Some of the methods marked by (*) in \autoref{tab:comp_sota} conducted their experiments on combined Bengali and Hindi datasets, but still our method shows better results than those. The empirical comparisons highlight the usefulness and reliability of the proposed method for signature verification.}

% \textcolor{black}{Our model performs significantly better than most of the works on the Bengali dataset and is marginally better than \cite{dey2017signet}. On the Hindi dataset, our model beats \cite{rateria2018off} by a very small margin in terms of accuracy, although it shines in its low FRR rate.} 

\begin{table}[!ht]
\centering
\caption{Comparison of our method with the SOTA. \blue{(Same results are presented for both the datasets for the methods marked by (*) as their experiments were conducted on combined Bengali and Hindi datasets).} \label{tab:comp_sota}}
\resizebox{\columnwidth}{!}{
\begin{tabular}{c|ccc|ccc} 
\hline
\multirow{2}{*}{\textbf{Method}} 
& \multicolumn{3}{c|}{\textbf{BHSig260 Bengali} \cite{pal2016performance}}               
& \multicolumn{3}{c}{\textbf{BHSig260 Hindi} \cite{pal2016performance}}                   
\\ \cline{2-7}
& \textbf{Accuracy (\%)} & \textbf{FAR (\%)} & \textbf{FRR (\%)} 
& \textbf{Accuracy (\%)} & \textbf{FAR (\%)} & \textbf{FRR (\%)}  
\\ \hline
% \blue{Auto-Encoder} \cite{baldi2012autoencoders}
% & \blue{83.57}      & \blue{29.11}      & \blue{3.75}          
% & \blue{89.14}      & \blue{18.47}      & \blue{3.25}           \\

\blue{SimCLR} \cite{chen2020simple}
& \blue{81.86}	    & \blue{30.22}	    & \blue{6.04}  
& \blue{66.49}	    & \blue{39.79}	    & \blue{27.21}          \\
\blue{Barlow Twins} \cite{zbontar2021barlow}
& \blue{82.75}	    & \blue{28.44}	    & \blue{6.25}
& \blue{72.20}	    & \blue{42.53}	    & \blue{13.06}          \\
\blue{SimSiam} \cite{chen2021exploring}
& \blue{82.61}	    & \blue{27.88}	    & \blue{6.87}	
& \blue{73.05}	    & \blue{29.16}	    & \blue{24.73}          \\
\cdashline{1-7}

Pal et al. \cite{pal2016performance}                       
& 66.18     & 33.82             & 33.82             
& 75.53     & 24.47             & 24.47              \\
Dey et al. \cite{dey2017signet}                       
& 86.11     & 13.89             & 13.89             
& 84.64     & 15.36             & 15.36              \\
Dutta et al. \cite{dutta2016compact}                    
& 84.90     & 15.78             & 14.43             
& 85.90     & 13.10             & 15.09              \\
Alaei et al. \cite{alaei2017efficient}*
& --        & 16.18             & 30.12
& --        & 16.18            & 30.12              \\
Rateria et al. \cite{rateria2018off}                   
& 75.06     & 27.81             & 21.74             
& 89.33     & 10.93             & 10.02              \\

\blue{Jadhav et al.} \cite{jadhav2018symbolic}*
& \blue{67.00}     & \blue{--}                & \blue{--} 
& \blue{67.00}     & \blue{--}                & \blue{--}                 \\
\blue{Bhunia et al.} \cite{bhunia2019signature}*
& \blue{--}        & \blue{24.10}             & \blue{26.00} 
& \blue{--}        & \blue{24.10}             & \blue{26.00}              \\
\blue{Jain et al.} \cite{jain2021signature}
& \blue{76.03}     & \blue{--}                & \blue{--}
& \blue{83.50}     & \blue{--}                & \blue{--}

\\ \hline
\textbf{Proposed Method}                 
& \textbf{87.34} & \textbf{19.89} & \textbf{5.42} 
& \textbf{89.50} & \textbf{12.01} & \textbf{8.98}    
\\ \hline
\end{tabular}
}
\end{table}

\subsection{Cross-Dataset Evaluation}\label{subsec:cross_dset}
Further, we extend our evaluations to a cross-dataset scenario i.e. training on one dataset and evaluating on a different dataset. The aim of this investigation is to find out the transferability of representations obtained from signatures of one language to another. To do so, We performed the experiments on two possible setups:

\begin{itemize}
    \item \textbf{Setup-1: } We train only the self-supervision pipeline on one dataset and perform the fine-tuning phase as well as evaluate on the other. %\textcolor{red}{This is much like a fine-tuning based transfer learning setup. Since the overall setup requires training samples from both datasets,} 
    We have maintained the train/test split as described in Section \ref{subsec:dataset}. The rest of the framework remains unchanged.
    
    \item \textbf{Setup-2(a): } We perform the entire training process i.e. self-supervised training followed by dual triplet loss-based fine-tuning, on one dataset. We then use this fine-tuned model for inference on a dataset containing signatures from a different language. For inference, the entire dataset was used.
    \blue{For this setup, we chose to follow the training set configuration as given in Section \ref{subsec:dataset} so that our training process remains consistent with those reported in Tables \ref{tab:comp_ablative} and \ref{tab:comp_sota}.}
    \blue {
    \item \textbf{Setup-2(b): } The setup is identical to Setup-2(a), except that we use the entire dataset for training.}
    
\end{itemize}

% Although we could have as well trained our model on an entire dataset for this setup, we nevertheless chose to follow the training set configuration as given in Section \ref{subsec:dataset} so that our training process remains consistent with those reported in Tables \ref{tab:comp_ablative} and \ref{tab:comp_sota}.

\begin{figure}[!hbt]
    \centering
    \subfloat[Evaluation on Bengali dataset]{\includegraphics[width=128pt, height=85pt]{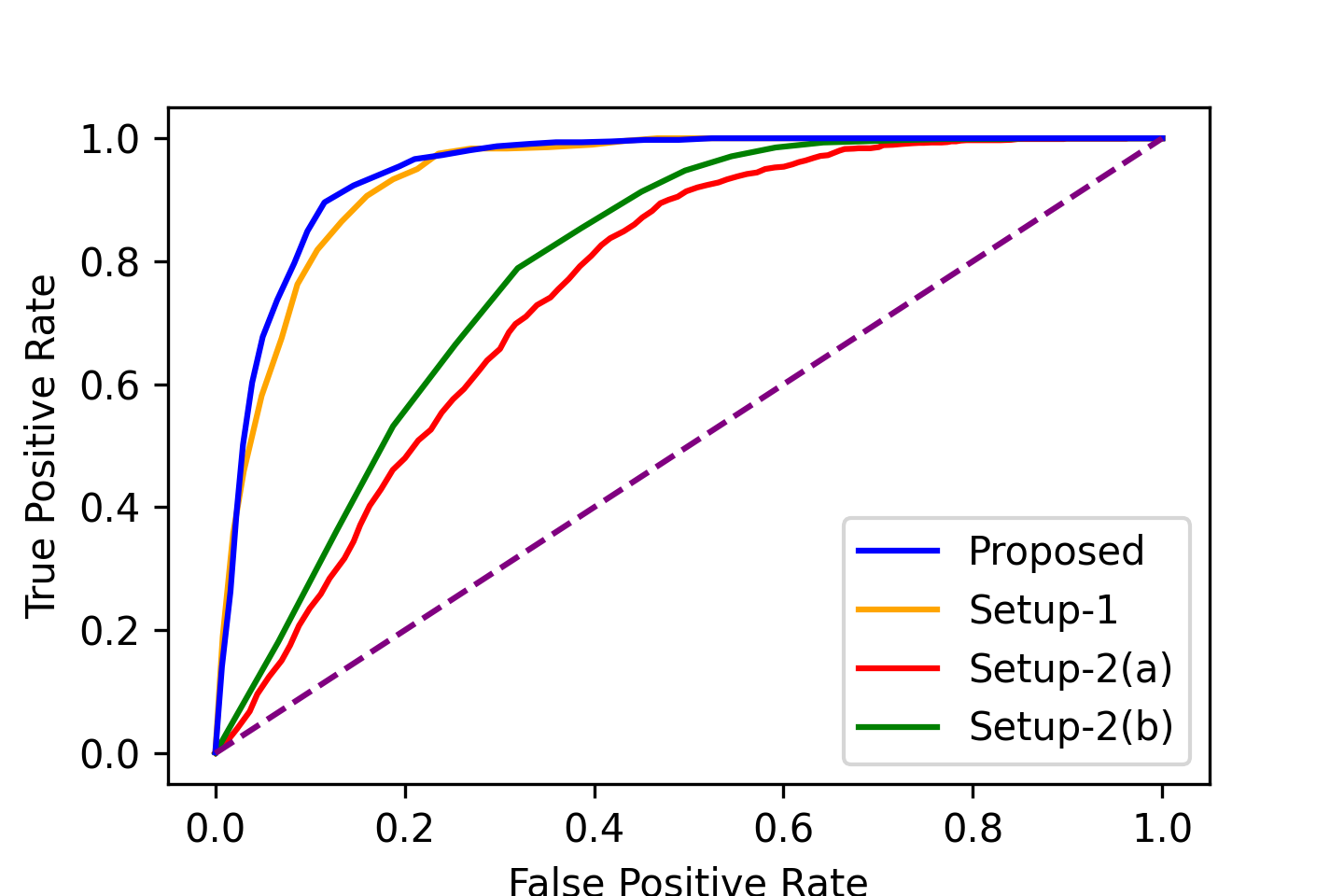}}
    \hspace{-2.5mm}
    \subfloat[Evaluation on Hindi dataset]{\includegraphics[width=128pt, height=85pt]{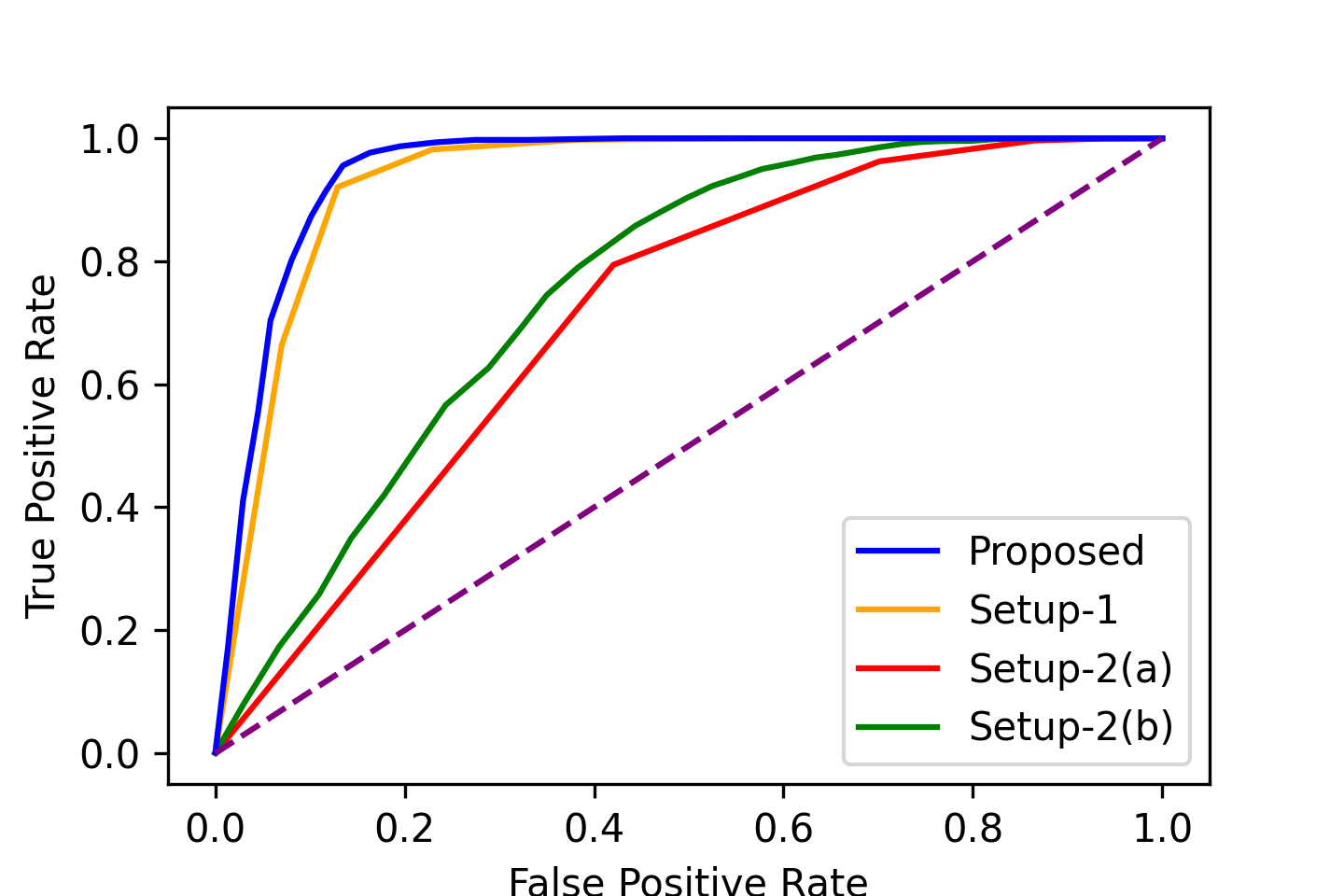}}
    \caption{\blue{ROC obtained across downstream experiments.}}
    \label{fig:roc}
\end{figure}

The results are shown in \autoref{tab:cross_dset}. \blue{The ROC curves presented in \autoref{fig:roc} depict the performance of the different setups of our proposed model.} We observe that in Setup-1 (i.e. fine-tuning based transfer learning setup), the results obtained are almost at par with state-of-the-art works as well as with the proposed method in an intra-dataset setup (tabulated in \autoref{tab:comp_sota}). For Setup-2(a), the performance detoriation is expected, since all datasets have some intrinsically unique features which can be learnt only from intra-dataset training. \blue{Further, it must also be noted that in Setup-2(a), we have not used the entire training dataset and instead used only a fraction of it to maintain consistency with our original experiments. The results show improvement when the entire training dataset is used in Setup-2(b).} Nevertheless, the cross-dataset transfer learning results for Setup-2(a,b) are empirically better than Dey et al. \cite{dey2017signet}, which reported accuracies of 64.57\% (train: Bengali, test: Hindi) and 60.65\% (train: Hindi, test: Bengali) on cross-dataset evaluation, which highlights the effectiveness of our proposed model to ably learn transferable representations across datasets. %signature representation learning process

%% Transpose table
\begin{table}[!ht]
\centering
\caption{Cross-dataset evaluation results. \label{tab:cross_dset}}
\resizebox{\columnwidth}{!}{
\begin{tabular}{c|ccc|ccc} 
\hline
\multirow{2}{*}{\textbf{Train/Test }} 
& \multicolumn{3}{c|}{\textbf{ Bengali/Hindi }}                  
& \multicolumn{3}{c}{\textbf{Hindi/Bengali }}                     \\ 
\cline{2-7}
& \textbf{Accuracy (\%)} & \textbf{FAR (\%)} & \textbf{FRR (\%)} 
& \textbf{Accuracy (\%)} & \textbf{FAR (\%)} & \textbf{FRR (\%)}  \\ 
\hline
Setup-1                               
& 88.19                  & 20.62             & 2.99              
& 86.14                  & 17.11             & 10.01              \\
Setup-2(a)                            
& 67.01                  & 51.14             & 14.82            
& 72.26                  & 46.46             & 9.00               \\
\blue{Setup-2(b)}                            
& \blue{69.59}           & \blue{41.87}      & \blue{14.65}             
& \blue{73.58}             & \blue{30.03}      & \blue{22.81}              \\
\hline
\end{tabular}
}
\end{table}

% ROC -- trial
% \vspace{-0.8cm}
% \vspace{-0.2cm}

%-------------------------------------------------------
% \begin{table}[!ht]
% \centering
% \caption{Cross-dataset evaluation results obtained by the proposed method. \label{tab:cross_dset}}
% \resizebox{\columnwidth}{!}{
% \begin{tabular}{c|ccc|ccc|ccc} 
% \hline
% \multicolumn{1}{c|}{\multirow{2}{*}{\textbf{Train/Test}}} 
% & \multicolumn{3}{c|}{\textbf{Setup-1}}                        
% & \multicolumn{3}{c|}{\textbf{Setup-2(a)}}  
% & \multicolumn{3}{c}{\textbf{\blue{Setup-2(b)}}} 
% \\ \cline{2-10}
% \multicolumn{1}{c|}{}                                      
% & \textbf{Accuracy (\%)} & \textbf{FAR (\%)} & \textbf{FRR (\%)} 
% & \textbf{Accuracy (\%)} & \textbf{FAR (\%)} & \textbf{FRR (\%)}  
% & \textbf{Accuracy (\%)} & \textbf{FAR (\%)} & \textbf{FRR (\%)} 
% \\ \hline
% Bengali/Hindi                                              
% & 88.19                  & 20.62             & 2.99              
% & 67.01                  & 51.14             & 14.82  
% & \blue{69.59}           & \blue{41.87}      & \blue{14.65}
% \\ \hline
% Hindi/Bengali                                              
% & 86.14                  & 17.11             & 10.01             
% & 72.26                  & 46.46             & 9.00    
% & \blue{73.58}             & \blue{30.03}      & \blue{22.81}
% \\ \hline
% \end{tabular}
% }
% \end{table}

\subsection{Cross-Script Evaluation}\label{subsec:cross_script}
Finally, we also probe into a cross-script OSV setup, where the model trained on a signature corpus of a given script is evaluated on a dataset of a different script. For this setup, we first train our proposed framework separately on the Bengali and Hindi datasets (i.e. Indic script) \cite{pal2016performance} and then test the trained models on the ICDAR-2011 Dutch signature dataset \cite{alvarez2016offline}. The Dutch dataset has a pre-defined evaluation set comprising query (1287) as well as reference (646) signatures from 54 writers. Here, we have adhered to the training set distribution described in Section \ref{subsec:dataset}. The results obtained on the Dutch dataset by training on the Indic scripts are:

\begin{itemize}
    \item \textbf{Bengali: } ACC: $68.60\%$, FAR: $43.51\%$, FRR: $19.29\%$
    \item \textbf{Hindi: } \hspace{4pt} ACC: $76.65\%$, FAR: $40.06\%$, FRR: $6.63\%$
\end{itemize}

It is worth noting that our setup is mere transfer learning without any explicit fine-tuning and thus, the model is completely unaware of any script specific knowledge. Yet, our model shows promise in the evaluation, especially with its low FRR values, which is commendable. The results show that our study has the potential to be explored in developing a script-independent universal signature verification framework.

\section{Conclusion and Future Work}\label{sec:concl}

In this study, we have proposed a novel two-stage pipeline for offline signature verification that employs a self-supervised reconstruction network for pre-training, which is then fine-tuned using a metric learning objective. The reconstruction model is ably augmented using a patch-wise 2D attention module which enforces the network to focus on local signature patches so as to learn meaningful representations that can be transferred to the downstream verification task. The fine-tuning phase uses a dual triplet objective that effectively discriminates between genuine and forged signature representations. We have shown by empirical results that our model shows promising performance on publicly available benchmark signature datasets, outperforming quite a few existing state-of-the-art works in literature. Further, comparisons with baseline competitors highlight the usefulness of the respective components of the proposed framework, and testing in challenging cross-dataset and cross-script setups emphasize its application in a diverse environment. Being first-of-its-kind, this work potentially paves the way for developing self-supervised pre-training tasks for the domain of signature verification, which is very important due to the bottleneck in availability of annotated data in the real world. Furthermore, the proposed pre-training pipeline may also be employed to other domains of computer vision, especially where region-wise focusing is crucial. We intend to work further in these directions in our future works.

% References
\bibliographystyle{IEEEtran}
\bibliography{IEEEfull,References}

\end{document}